\documentclass[runningheads]{llncs}

\usepackage{eccv}

\usepackage{eccvabbrv}

\usepackage{graphicx}
\usepackage{booktabs}

\usepackage{wrapfig}
\usepackage{bbding}
\usepackage{floatrow}
\newfloatcommand{capbtabbox}{table}[][\FBwidth]
\usepackage{xcolor}
\usepackage{listings}
\usepackage{color}
\usepackage{colortbl}
\usepackage{amsmath}
\usepackage{amssymb}
\usepackage{multirow}
\usepackage{enumitem}  
\usepackage{lipsum}
\usepackage[misc]{ifsym} 
\usepackage[accsupp]{axessibility}  

\usepackage[pagebackref,breaklinks,colorlinks,citecolor=eccvblue]{hyperref}

\usepackage{orcidlink}

\begin{document}

\title{{Q\&A Prompts}: Discovering Rich Visual Clues through Mining Question-Answer Prompts for VQA requiring Diverse World Knowledge} 

\titlerunning{{Q\&A Prompts}}

\author{Haibo Wang\orcidlink{0009-0008-4690-3375} \and Weifeng Ge$^{\textrm{\Letter}}$\orcidlink{0000-0002-6258-6225}}


\authorrunning{Haibo Wang \and Weifeng Ge}

\institute{
School of Computer Science,\ Fudan University\\
\email{hbwang22@m.fudan.edu.cn, wfge@fudan.edu.cn}\\
}

\maketitle

\newcommand\blfootnote[1]{%
\begingroup
\renewcommand\thefootnote{}\footnote{#1}%
\addtocounter{footnote}{-1}%
\endgroup
}


\begin{abstract}
With the breakthrough of multi-modal large language models (MLLMs), answering complex visual questions that demand advanced reasoning abilities and world knowledge has become a much more important testbed for developing AI models than ever. However, equipping MLLMs with robust cross-modality reasoning ability remains challenging since the cognition scheme of humans has not been understood systematically. In this paper, we believe that if we can collect rich visual clues, we will recognize the image more accurately, understand the question better, recall relevant knowledge more easily, and finally reason out the answer. We discover these rich visual clues by mining question-answer pairs in images and sending them into multi-modal large language models as prompts. We call the proposed method {Q\&A Prompts}. Specifically, we first use the image-answer pairs and the corresponding questions in the training set as inputs and outputs to train a visual question generation (VQG) model. Then, we use an image tagging model to identify various instances and send packaged image-tag pairs into the VQG model to generate relevant questions with the extracted image tags as answers. Finally, we encode these generated question-answer pairs as prompts with a visual-aware prompting module and send them into pre-trained MLLMs to reason out the final answers. Experimental results show that, compared with state-of-the-art methods, our {Q\&A Prompts} achieves substantial improvements on the challenging visual question answering datasets requiring reasoning over diverse world knowledge, such as OK-VQA and A-OKVQA. Codes will be avaliable at \href{https://github.com/WHB139426/QA-Prompts-ECCV-24}{link}.
\end{abstract}

\section{Introduction}
\label{sec:intro}

\begin{figure}[t]
\label{fig:fig1}
  \centering
   \includegraphics[scale=0.26]{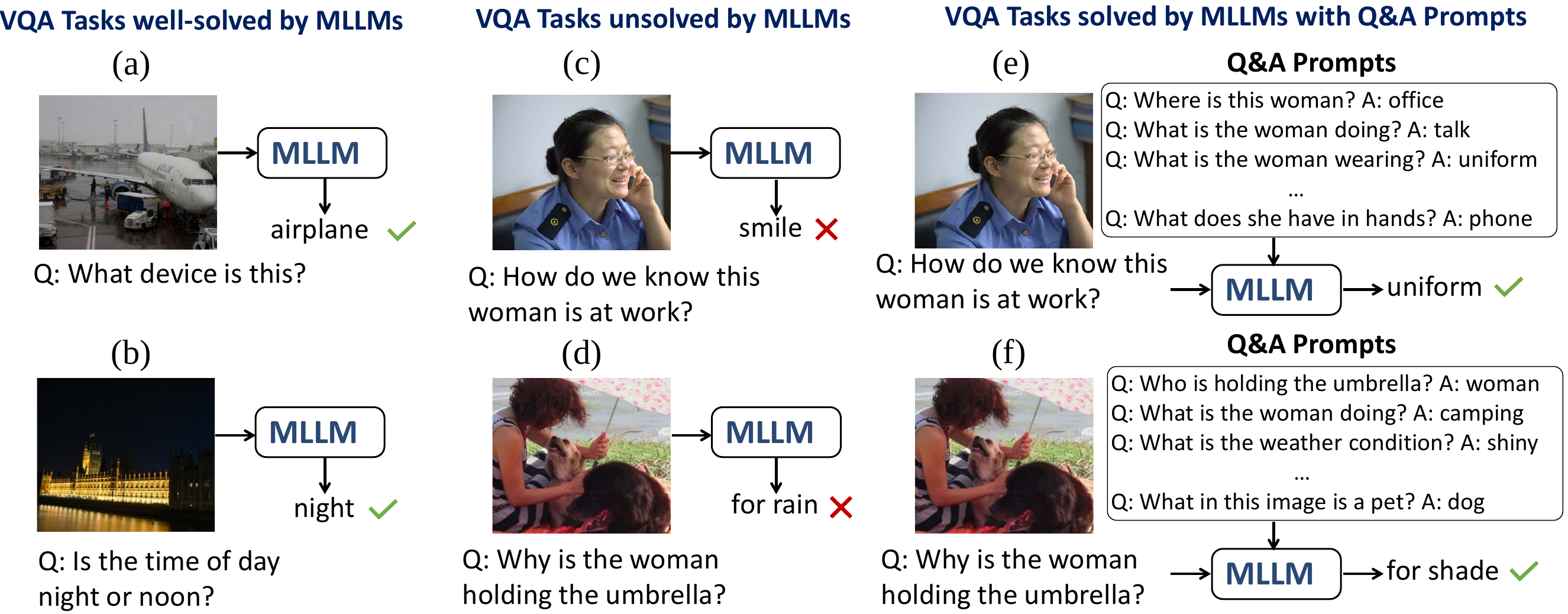}
   \caption{Illustration of VQA tasks that require simple perception abilities ((a) and (b)) and those demanding chains of reasoning over diverse world knowledge ((c) and (d)). The example on the right ((e) and (f)) indicates MLLMs with Q\&A prompts can solve difficult VQA problems that require reasoning over diverse world knowledge.}
   \label{fig:fig1}
\end{figure}

Visual Question Answering (VQA) has longstanding been considered a Visual Turing Test~\cite{turing} in the machine learning community. To solve this Turing test, artificial intelligence (AI) models are expected to be equipped with the human capabilities of visual recognition, language understanding, logical reasoning, world knowledge referring, etc. There are a variety of datasets focusing on different problems in VQA, such as the perception and language understanding problem \cite{vqa_2015,vqa_2017}, reasoning on procedurally generated images \cite{clevr}, and structured factual knowledge \cite{wang2015explicit,wang2017fvqa} or commonsense knowledge \cite{vcr}. Recently, with the advent of multi-modal large language models (MLLMs) \cite{blip2, instruct_blip, mplug, llava}, significant progress has been achieved in perception-based VQA tasks \cite{vqa_2015, vqa_2017}. \cref{fig:fig1} (a) and \cref{fig:fig1} (b) show such VQA tasks, which can be reliably solved by MLLMs. 

However, for more complex VQA tasks \cite{okvqa,aokvqa} as in \cref{fig:fig1} (c) and \cref{fig:fig1} (d), that feature the necessity of diverse world knowledge and complex chains of reasoning, state-of-the-art methods \cite{kat, revive, reveal} even MLLM models \cite{blip2, instruct_blip} fail to give the correct answers. Different from perception-based VQA \cite{vqa_2015, vqa_2017} and "closed" knowledge-based VQA \cite{wang2015explicit, wang2017fvqa}, visual question reasoning in OK-VQA \cite{okvqa} and A-OKVQA \cite{aokvqa} requires not only diverse forms of world knowledge but also need complex reasoning logic. In \cref{fig:fig1} (d), where a woman sits on the ground with her pet dogs and holds an umbrella, the question is [{\ttfamily{Why is the woman holding the umbrella?}}]. Simply associating the keyword [{\ttfamily{umbrella}}] with the word [{\ttfamily{rain}}] due to their frequent co-occurrence in the pre-training corpus \cite{vqa_biases} will lead to the wrong answer. To reason out the correct answer, AI models need to consider rich visual clues presented in the image. For example, in \cref{fig:fig1} (f), having discovered the clues including [{\ttfamily{the woman is camping}}] and [{\ttfamily{the weather is shiny}}] from the Q\&A pairs, the model can correctly fill the logic gap and arrive at the answer [{\ttfamily{for shade}}]. This evidence indicates the necessity of collecting rich visual clues to achieve good results.

Based on the analysis above, we believe that collecting visual clues of various instances in images will help MLLM recognize the image more accurately, understand the question better, recall relevant knowledge more easily, and finally generate the reasoning results more intelligently. To investigate this hypothesis, we design a novel VQA framework called {\textit{Q\&A Prompts}}, which extracts numerous question-answer pairs associated with different perspectives in images as prompts for MLLMs. We choose an instance in the given image as the answer and generate the related question to form a question-answer pair. We generate the question-answer pairs for almost every instance to get a bag of Q\&A prompts. The reason we choose question-answer pairs as prompts comes from two folds: First, Q\&A prompts can focus on diverse instances in images like objects, scenes, or actions, thereby offering multiple different perspectives to understand the given image and the target question; Second, the knowledge and reasoning insights hidden in these question-answer pairs can help to explicitly uncover more relevant world knowledge, which may be critical to filling the logic gap between perception and reasoning when answering the target question. 

In {\textit{Q\&A Prompts}}, there are three key stages, which are the visual question generation (VQG) model training stage, the question-answer prompts generation stage, and the visual-language reasoning stage. In the VQG model training stage, we gather question-answer pairs in a VQA training set to train a  VQG model that can map an answer together with the given image to a correlated question. Subsequently, in the question-answer prompts generation stage, we leverage a pre-trained image tagging model \cite{ram, tag2text} to recognize various objects, scenes, and actions in images with text tags. We employ these tags as answers and send them into the VQG model along with the corresponding images to generate questions. Finally, in the visual-language reasoning stage, we encode the generated question-answer pairs with a visual-aware prompting module and send them together with image and question features into a frozen language model to reason out the answer. Since previous VQA benchmarks ~\cite{vqa_2015, vqa_2017} have seen saturated performance, we conduct experiments on the more challenging OK-VQA \cite{okvqa} and A-OKVQA \cite{aokvqa} datasets, which are newer and larger testbeds for MLLM to investigate their abilities in accomplishing reasoning tasks requiring diverse world knowledge. Experimental results show that our {\textit{Q\&A Prompts}} can substantially improve the reasoning ability of MLLMs, such as InstructBLIP \cite{instruct_blip}, LLaVA \cite{llava}, and MiniGPT-4 \cite{minigpt4}. In summary, we make the following contributions: 

\begin{itemize}
\item We propose a novel VQA framework that effectively generates and leverages question-answer pairs as prompts to improve the reasoning ability of multi-modal large language models. It provides a new perspective for the community to design VQA solutions by explicitly collecting rich visual clues to bridge the logic gap between perception and reasoning when answering the target question.
\item We design a novel question-answer prompts generation scheme with a VQG model and an image tagging model, which can generate Q\&A prompts for recognizable objects, scenes, and activities in images. Besides, we design a new visual-aware prompting module to encode these prompts efficiently for the subsequent reasoning. 
\item We test the method on the challenging OK-VQA \cite{okvqa} and A-OKVQA \cite{aokvqa} benchmarks, which study reasoning over diverse forms of world knowledge. The proposed {\textit{Q\&A Prompts}} achieves an accuracy of \textbf{68.1$\%$} and \textbf{64.3$\%$} on A-OKVQA \cite{aokvqa} and OK-VQA \cite{okvqa}, outperforming previous state-of-the-art methods by clear margins.
\end{itemize}


\section{Related Work}
\label{sec:related_work}
\textbf{Multimodal Large Language Models (MLLMs).} MLLMs \cite{blip2, falmingo, instruct_blip, lavin, llama_adapter, minigpt4, mplug, otter} have shown strong ability in image-language understanding and reasoning, by adapting frozen language models to frozen image encoders with trainable connection modules. For instance, Flamingo \cite{falmingo} incorporates visual features into the language model with gated cross-attention blocks. LLaVA \cite{llava}, instead, directly projects visual features into the space of text embeddings using a linear layer. BLIP-2 and InstructBLIP \cite{blip2, instruct_blip} introduce a more complex Q-Former to bridge the modality gap. These models align images and language based on large-scale image-text pre-training \cite{laion2b, coco, cc12m} and conduct reasoning with language models. In this paper, we mine more numerous question-answer pairs and send them into the MLLM as prompts. These prompts provide rich visual clues and uncover relevant world knowledge, which is beneficial to reasoning out correct answers.  

\noindent \textbf{VQA Requiring Reasoning over Diverse World Knowledge.} 
Different from classical VQA tasks \cite{vqa_2015, vqa_2017,wang2015explicit,wang2017fvqa}, VQA requiring reasoning over world knowledge involves inferring intents, goals, the physics of the world, and the social dynamics of individuals. More recently, A-OKVQA \cite{aokvqa} and OK-VQA \cite{okvqa}, which our work focuses on, present large-scale challenging VQA datasets that encompass questions requiring reasoning based on diverse world knowledge, including visual knowledge, commonsense knowledge, and factoid knowledge. Previous methods like Img2LLM \cite{img2prompt}, Prophet \cite{prophet}, and PromptCap \cite{promptcap} exploit the rich hidden knowledge and strong reasoning abilities of LLMs (e.g., GPT3 \cite{gpt3}) to solve such problems. These methods convert images into natural languages and construct various prompt templates for LLMs with in-context learning \cite{gpt3} to generate answers. Another group of works, such as KAT \cite{kat}, REVIVE \cite{revive}, and REVEAL \cite{reveal}, directly encode questions and relevant knowledge retrieved from external knowledge bases \cite{wikidata, concepetnet, gpt3} to train end-to-end models. Unlike these methods, we design a visual-question prompts generation scheme to mine the information in images and unambiguous words to reduce confusion when answering the target questions. With Q\&A prompts, different types of world knowledge are incorporated into the reasoning process to help MLLMs infer the correct answers.

\noindent  \textbf{Visual Question Generation (VQG).} Various works have explored Visual Question Generation (VQG) in the context of VQA tasks.  SQuINT \cite{vqg_squinting} and SelTDA \cite{vqg_augment} focus on dataset augmentation to enhance the generalization of VQA models. SQuINT employed manual workers to annotate questions, while SelTDA generated questions automatically using unlabeled images. Another line of research, exemplified by \cite{vqg_informative}, employs a VQG model to generate informative sub-questions. These sub-questions are then used alongside the original question during inference, enabling another VQA model to answer both simultaneously. Different from previous works, our VQG model uses the tags of images \cite{ram, tag2text} as answers for generated questions. This approach ensures that our VQG model generates more relevant questions with additional details and perspectives. Furthermore, we introduce a novel module to encode the question-answer pairs as prompts for MLLMs, instead of directly feeding them as the formation of natural language into the model \cite{vqg_informative}.

\begin{figure*}[t]
\label{fig:fig2}
  \centering
   \includegraphics[scale=0.307]{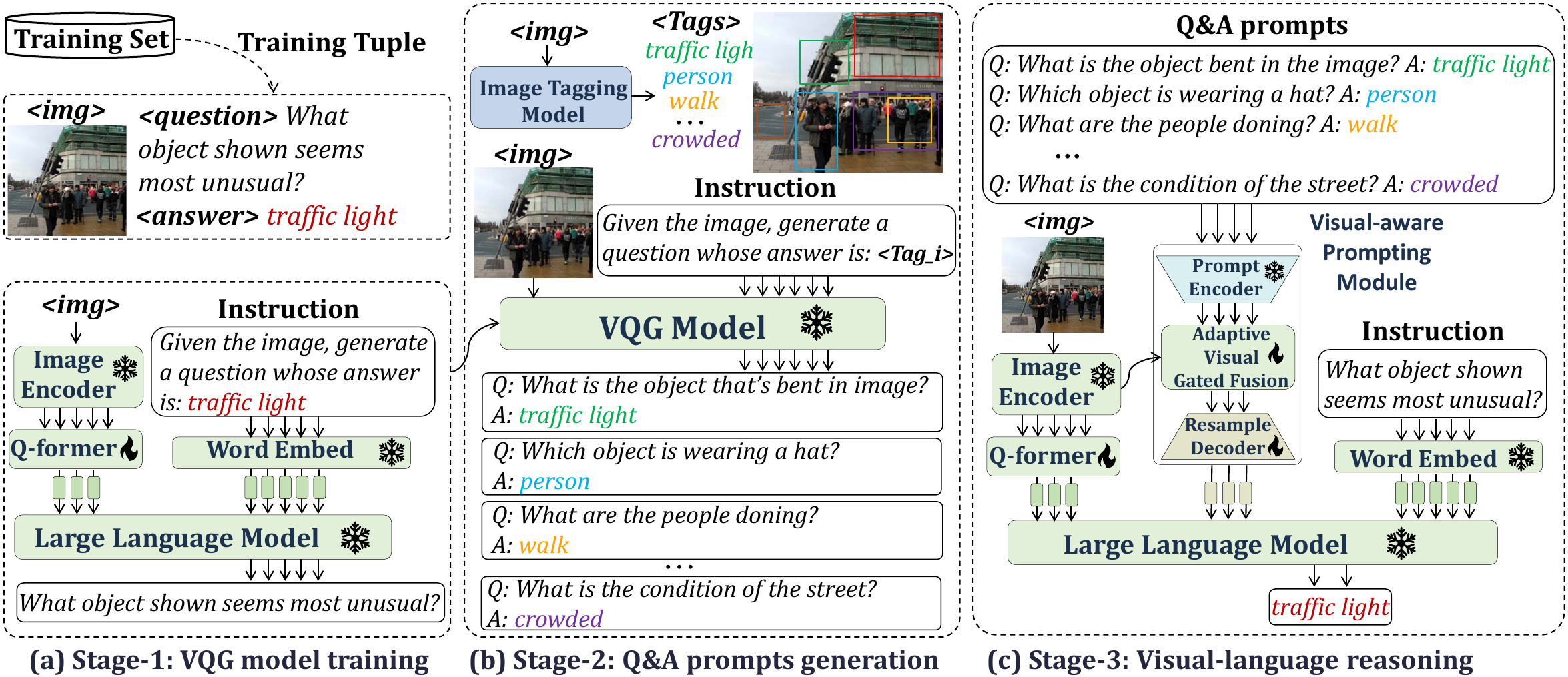}
   \caption{(a) Firstly, we train a VQG model that can ask informative questions given the image and specified answer. (b) Then, we exploit the image tagging model to extract image tags as the answers to generated questions and obtain diverse question-answer prompts. (c) At last, we feed the image, question, and question-answer prompts into the VQA model to perform visual-language reasoning in (c).}
   \label{fig:fig2}
\end{figure*}

\section{Method}
\label{sec:method}

{\textit{Q\&A Prompts}} is a conceptually simple three-stage framework, and \cref{fig:fig2} gives an overview: a VQG model training stage, a question-answer prompts generation stage, and a visual-language reasoning stage. Specifically, in the VQG model training stage (\cref{fig:fig2} (a)), we train a vanilla MLLM with a broad range of image-answer pairs as input and the corresponding question as the target output to learn a mapping from answers to questions (detailed in \cref{vqg}). In the prompts generation stage (\cref{fig:fig2} (b)), we exploit an image tagging model \cite{tag2text, ram} to obtain relevant tags including attributes, objects, actions, and scenes in images. These tags serve as answers and will be fed into the VQG model trained in the first stage to generate diverse questions containing rich visual clues and world knowledge (detailed in \cref{tag}). In the final reasoning stage (\cref{fig:fig2} (c)), we utilize our proposed visual-aware prompting module to encode these question-answer pairs into prompts, which will be fed into the MLLM such as BLIP-2 \cite{blip2} or InstructBLIP \cite{instruct_blip} together with the image and the target question, to predict the answer (detailed in \cref{reason}). The whole pipeline is flexible and effective in mining informative visual clues to correctly answer the given questions.

\subsection{Stage-1: Train the VQG Model}
\label{vqg}
Given an image $\mathcal{V}$ and answer $\mathcal{A}$, the VQG model is responsible for posing an informative question $\mathcal{Q}$ which can be properly answered by $\mathcal{A}$. To explore as many visual clues in the generated questions as possible, and avoid generating general questions containing limited clues (e.g., "What is this?"), we utilize the A-OKVQA \cite{aokvqa} or OK-VQA \cite{okvqa} as the training set $D$ for our VQG model, since the questions in their annotations are diverse and specific, covering broader areas of knowledge. To create the VQG model that approximates $P(\mathcal{Q}|\mathcal{V}, \mathcal{A})$, we treat the problem of learning such a model as a text-generation problem and wish to train the MLLM (e.g., InstructBLIP \cite{instruct_blip}) as our VQG model with $D$ due to its strong image-language reasoning ability.

Specifically, as shown in \cref{fig:fig2} (a), the VQG model comprises a frozen image encoder, a trainable connection module, and a frozen LLM. We extract image embeddings $\textbf{E}_{v} \in \mathbb{R}^{n\times d_v}$ of $\mathcal{V}$ with the image encoder ($n$ is the patch number, $d_v$ is the embedding dimension), and then feed $\textbf{E}_{v}$ into the connection module to obtain fixed-length visual features $\mathbf{F}_v \in \mathbb{R}^{k\times d_q}$, which will be sent into the LLM together with the instruction. We write the instruction as the template containing the answer $\mathcal{A}$, such as \textit{"Given the image, generate a question whose answer is: $\langle \mathcal{A}\rangle$."}. The model is trained using the cross-entropy loss with parameters $\theta$:
\begin{equation}
\begin{aligned}
  \mathcal{L_{VQG}} = -\sum_{t=1}^{L_{q}}logP_{\theta}(\mathcal{Q}_t|\mathcal{Q}_{<t}, \mathcal{V}, \mathcal{A})
  \label{eq:loss_vqg}
\end{aligned}
\end{equation}
where $\mathcal{Q}_t$ is predicted autoregressively at position $t$, and $L_q$ is the sequence length of the question $\mathcal{Q}$. We emphasize the necessity of training the VQG model, instead of prompting a frozen MLLM in a zero-shot manner. In \cref{fig:fig_vqg}, we use sentence-bert \cite{sentence-bert} to obtain the dense embedding vectors of each question and show the T-SNE \cite{tsne} embedding of them, as well as visualization examples. We can observe that the questions generated by our VQG model (\textcolor{cyan}{blue}) are more diverse and specific like questions in A-OKVQA (\textcolor{teal}{green}), while the questions generated by MLLM without training (\textcolor{orange}{orange}) are often general and limited with low differentiation. Our following experiments will further verify this.

\begin{figure}[t]
\label{fig:fig_vqg}
  \centering
   \includegraphics[scale=0.4]{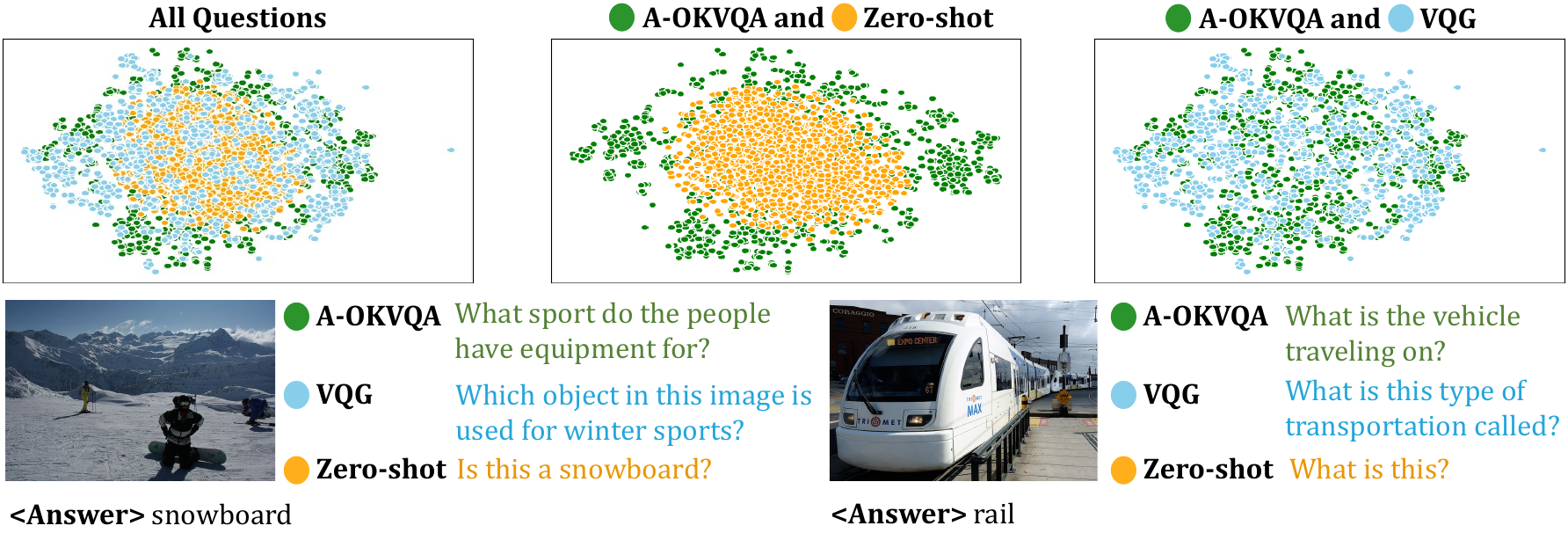}
   \caption{Visualizations and T-SNE embedding distributions of the questions in manually annotated A-OKVQA (\textcolor{teal}{green}), and the questions generated by our VQG model (\textcolor{cyan}{blue}) or by directly prompting a frozen MLLM (\textcolor{orange}{orange}).}
   \label{fig:fig_vqg}
\end{figure}

\subsection{Stage-2: Generate Question-Answer Prompts}
\label{tag}
Once the VQG model has been obtained, the generation of question-answer pairs can proceed. In this stage, we will explicitly generate a wide range of question-answer pairs to help the model have a deeper and broader understanding of the image. In order to generate diverse questions covering various perspectives of the image, we should ensure the answers fed into the VQG model are sufficient in varied forms. Therefore, we adopt a strong image tagging model \cite{tag2text, ram} to produce rich tags as answers. Different from object detectors that can only recognize limited object labels, image tagging models aim to identify various open-vocabulary elements in images, including attributes, objects, actions, scenes, etc, which can provide an appropriate source of the answers we need.

We utilize the RAM \cite{ram} model as our image tagging model due to its superior recognition ability. As shown in \cref{fig:fig2} (b), we feed the image $\mathcal{V}$ into the tagging model and obtain various tags for each image. To make the tags of each image more distinguishable, we remove the most frequent ones containing general information, such as 'person', 'food', 'man', etc., resulting in an average of 14 tags per image. We denote this set of tags as $\mathcal{O}=\{o_i\}_{i=1}^{M}$, where $M$ is the number of tags associated with an image.

Using these tags as answers, we input them into the trained VQG model along with the image, as shown in \cref{fig:fig2} (b), to obtain corresponding questions that answer these tags. To ensure that the generated questions (denoted as $\mathcal{T}=\{t_i\}_{i=1}^M$ ) contribute effectively to answering the target question, we rank the generated questions $\mathcal{T}=\{t_i\}_{i=1}^M$ based on their similarity with the target question $\mathcal{Q}$, and adopt the questions with the $\operatorname{Top-P}$ similarities as the final obtained question prompts $\mathcal{U}$:
 \begin{equation}
  \mathcal{U} = \{u_p\}_{p=1}^{P} = \mathop{\arg\text{Top-P}}\limits_{t_i\in \mathcal{T}}{\langle \operatorname{T}(t_i),\operatorname{T}(\mathcal{Q})\rangle}
  \label{eq:similarity}
\end{equation}
where $\operatorname{T}(\cdot)$ represents the embedding model \cite{sentence-bert}, which encodes sentences into embedding vectors for similarity computation. The $\langle \cdot,\cdot\rangle$ operator is the inner product, computing the cosine similarity between two sentences. We then concatenate each $u_i$ in $\mathcal{U}$ and its corresponding $o_i$ in $\mathcal{O}$ using the template \textit{"Question: $\langle u_i\rangle$ Answer: $\langle o_i\rangle$"}, resulting in the final question-answer pairs as prompts $\mathcal{S}$.

\begin{figure}[t]
\label{fig:vpm}
  \centering
   \includegraphics[scale=0.65]{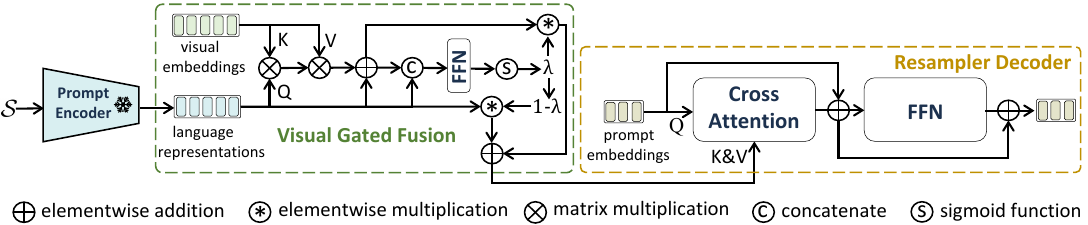}
   \caption{Overview of the visual-aware prompting module, consisting of the prompt encoder, visual gated fusion, and resampler decoder.}
   \label{fig:vpm}
\end{figure}

\subsection{Stage-3: Reasoning with Q\&A Prompts}
\label{reason}
After obtaining these question-answer pairs involving rich visual clues and world knowledge, we use the proposed visual-aware prompting module, as shown in \cref{fig:vpm}, to encode these question-answer pairs into a set of prompt embeddings $\textbf{F}_p \in \mathbb R^{k\times d_{q}}$ for our VQA model. We still utilize the same MLLM (e.g., InstructBLIP \cite{instruct_blip}) as our VQA model for the final reasoning. As in \cref{fig:fig2} (c), the encoded prompt embeddings $\textbf{F}_p$ will be fed into the frozen LLM in MLLM, together with the instruction embeddings and visual features $\mathbf{F}_v$ to predict the answer $\mathcal{A}$:
 \begin{equation}
 \begin{aligned}
  \mathcal{A} = &\operatorname{LLM}([\textbf{F}_v;\textbf{F}_p;\operatorname{Embed}(\mathrm{Ins})])
  \label{eq:llm_decoder_a}
  \end{aligned}
\end{equation}
$\operatorname{Embed}$ is the word embedding layer in LLM, the operator $[;]$ means concatenation, and $\mathrm{Ins}$ is the instruction including the target question $\mathcal{Q}$, represented as \textit{"Question: $\langle \mathcal{Q}\rangle$? Answer: "}. We then introduce how the visual-aware prompting module bootstrapped the reasoning ability of MLLM within {\textit{Q\&A Prompts}}.

\noindent \textbf{Prompt Encoder.} The prompt encoder is responsible for obtaining the semantics representations of the prompts $\mathcal{S}$. Specifically, we choose encoder-based language models as our prompt encoder due to their strong ability in language understanding. To preserve the well-trained representations, we keep the parameters of the prompt encoder frozen. We feed the prompts $\mathcal{S}$ into the prompt encoder and get the encoded $\textbf{F}_{s}\in \mathbb R^{L_{s}\times d_{q}}$, where $L_{s}$ is the sequence length of question-answer prompts and $d_{q}$ is the hidden size of the prompt encoder.

\noindent \textbf{Visual Gated Fusion.} Since the output $\textbf{F}_{s}$ from the prompt encoder is an unimodal language representation, to fully utilize the explicit information in images, the visual gated fusion adaptively incorporates image information into this representation with the visual embeddings $\textbf{E}_v\in \mathbb{R}^{n\times d_v}$. Specifically, in \cref{fig:vpm}, we use a single-head attention mechanism, similar to MM-COT \cite{mmcot}, to correlate the language representation with the visual embeddings. We define the query, key and value as $\textbf{Q} = \textbf{F}_{s}\in \mathbb R^{L_{s}\times d_{q}}$, $\textbf{K} = \textbf{V} = \operatorname{MLP}(\textbf{E}_v)\in \mathbb R^{n\times d_{q}}$, where $\operatorname{MLP}$ here is projecting $d_v$ to $d_q$. The attention output $\textbf{F}^{attn}_{v}$ is defined as:
\begin{equation}
  \textbf{F}^{attn}_{v} = \textbf{Q} + \operatorname{Softmax}(\frac{\textbf{Q}\textbf{K}^{T}}{\sqrt{d_q}})\textbf{V},\quad \textbf{F}^{attn}_{v}\in \mathbb R^{L_{s}\times d_{q}}
  \label{eq:qkv_1}
\end{equation}
Next, we apply the gated fusion mechanism to merge $\textbf{F}_{s}$ and $\textbf{F}^{attn}_{v}$. The fused representation $\textbf{F}_{m}\in \mathbb{R}^{L_{s}\times d_{q}}$ is obtained using the following equations:
\begin{equation}
\begin{aligned}
  \lambda &= \operatorname{Sigmoid}(\textbf{F}_{s}\textbf{W}_{s} + \textbf{F}^{attn}_{v}\textbf{W}_{v})
\\ &\textbf{F}_{m} = (1-\lambda)\textbf{F}_{s} + \lambda \textbf{F}^{attn}_{v}
  \label{eq:qkv_2}
\end{aligned}
\end{equation}
in which $\textbf{W}_{s}$ and $\textbf{W}_{v}$ are learnable parameters. The gated fusion allows the model to adaptively blend the information from the language representation and visual embeddings, resulting in the multimodal representations $\textbf{F}_{m}$.

\noindent \textbf{Resampler Decoder.}
\label{reason_cross}
Although $\textbf{F}_{m}\in \mathbb{R}^{L_{s}\times d_{q}}$ in \cref{eq:qkv_2} is the multimodal representation of $\mathcal{S}$, we cannot directly concatenate it with $\textbf{F}_{v}$ and instruction embeddings as in \cref{eq:llm_decoder_a}, due to the long sequence length of $L_{s}$, which can result in extra computational burden and distraction to the target question $\mathcal{Q}$. In light of this, we employ a lightweight resampler decoder inspired by Perceiver \cite{perceiver_io}, to map $\textbf{F}_{m}\in \mathbb{R}^{L_{s}\times d_{q}}$ into smaller-sized prompt embeddings $\textbf{F}_{p}\in \mathbb{R}^{k\times d_{q}}$ with fixed length $k$, where $k \ll L_{s}$ (e.g., $k=32$ and $L_{s}\geq 100$).

Concretely, we additionally introduce the learnable embeddings $\textbf{F}_{p}\in \mathbb R^{k\times d_{q}}$ to interact with $\textbf{F}_{m}$. As in \cref{fig:vpm}, a cross-attention layer is applied by taking $\textbf{F}_{p}$ as query, and $\textbf{F}_{m}$ as key and value, followed with a feed-forward network, to generate prompt embeddings with the multimodal representations from $\textbf{F}_{m}$:
\begin{equation}
\begin{aligned}
  \textbf{F}_{p} = \operatorname{Cross-Attention}(\textbf{F}_{p}, \textbf{F}_{m}) + \textbf{F}_{p}\\
  \textbf{F}_{p} = \operatorname{FFN}(\textbf{F}_{p}) + \textbf{F}_{p}, \ \textbf{F}_{p} = \operatorname{MLP}(\textbf{F}_{p})
  \label{eq:cross_attention} 
\end{aligned}
\end{equation}
where $\operatorname{MLP}$ projects $d_q$ to $d_{lm}$ (the hidden size of LLM). 

At last, the final output of $\textbf{F}_{p} \in \mathbb{R}^{k\times d_{q}}$ will be fed into the LLM together with the question instruction embeddings and visual features as in \cref{eq:llm_decoder_a}, to make predictions of answers. The model is trained using the cross-entropy loss function with trainable parameters $\theta$:
\begin{equation}
\begin{aligned}
  \mathcal{L_{VQA}} = -\sum_{t=1}^{L_{a}}logP_{\theta}(\mathcal{A}_t|\mathcal{A}_{<t}, \mathcal{V}, \mathcal{Q}, \mathcal{S})
  \label{eq:loss_ans}
\end{aligned}
\end{equation}
where $\mathcal{A}_t$ is predicted autoregressively at position $t$, and $L_a$ is the sequence length of the ground truth answer text $\mathcal{A}$.

\section{Experiments}
\label{sec:experiments}

\subsection{Dataset}
\label{dataset}
We evaluate \textit{Q\&A Prompts} on the challenging A-OKVQA \cite{aokvqa} and OK-VQA \cite{okvqa} since they are currently the largest VQA datasets requiring complex reasoning over diverse world knowledge. Specifically, the A-OKVQA dataset consists of 24,903 samples, with 17.1k samples for training, 1.1k for validation, and 6.7k for testing. The questions in this dataset require reasoning over various types of world knowledge, such as commonsense knowledge, factoid knowledge, and visual knowledge. The OK-VQA dataset includes 14k questions covering a variety of knowledge categories, with 9k samples for training, and 5k for validation. 
Each sample in these two datasets includes an image, a question, and 10 ground-truth answers. We use the soft accuracy \cite{vqa_2015} as the standard evaluation metric.

\begin{table}
  \centering
    \resizebox{\linewidth}{!}{
  \begin{tabular}{l|cc|c|cc}
    \toprule
    \multirow{2}{*}{Method} & \multirow{2}{*}{Image Representation} & \multirow{2}{*}{Knowledge Source}  & \multirow{2}{*}{OK-VQA (\%)} & \multicolumn{2}{c}{A-OKVQA (\%)}\\
    \cmidrule(lr){5-6}
     &    &   &   & Val & Test\\
    \midrule 
    ClipCap \cite{clipcap}             & Feature & Pretrain                   & -    & 18.1 & 15.8 \\
    Pythia \cite{pythia}               & Feature & Pretrain                   & -    & 25.2 & 21.9 \\
    ViLBERT \cite{vilbert}             & Feature & Pretrain                   & -    & 30.6 & 25.9\\
    LXMERT \cite{lxmert}               & Feature & Pretrain                   & -    & 30.7 & 25.9 \\
    GPV-2 \cite{gpv2}                  & Feature & Pretrain                   & -    & 48.6 & 40.7 \\
    Unified-IO \cite{unifiedio}        & Feature & Pretrain                   & 54.0 & -    & 45.2 \\
    Flamingo (80B) \cite{falmingo}     & Feature & Pretrain                   & 57.8 & -    & -    \\
    BLIP-2 \cite{blip2}            & Feature & Pretrain                   & 59.3 & 60.0 & 58.7 \\
    \midrule 
    Mucko  \cite{mucko}                & Feature & DBPedia + ConceptNet       & 29.2 & -    & -    \\
    ConceptBERT  \cite{conceptbert}    & Feature & NumberBatch + ConceptNet   & 33.7 & -    & -    \\
    KRISP \cite{krisp}                 & Feature & Wikipedia + ConceptNet     & 38.9 & 33.7 & 22.1 \\
    MAVEx \cite{mavex}                 & Feature & Wikipedia + ConceptNet     & 40.3 & -    & -    \\
    UnifER \cite{unifer}               & Feature & ConceptNet                 & 42.1 & -    & -    \\
    TRiG \cite{trig}                   & Caption + Tags + OCR & Wikipedia     & 49.4 & -    & -    \\
    REVEAL \cite{reveal}               & Feature & WIT + Wikidata             & 59.1 & -    & 52.2 \\
    RA-VQA-v2 \cite{ravqa}             & Feature & Google Search              & 62.1 & -    & -    \\
    \midrule 
    PICa \cite{pica}                   & Caption + Tags & Frozen GPT-3 (175B) & 48.0 & -    & -    \\
    KAT \cite{kat}                     & Caption + Feature & Wikidata + GPT3 (175B)  & 53.1 & -    & -    \\
    Img2LLM \cite{img2prompt}          & Caption & Frozen GPT-3 (175B)        & - & 42.9 & 40.7 \\
    REVIVE \cite{revive}               & Caption + Feature & Wikidata + GPT3 (175B)  & 56.6 & -    & -    \\
    PromptCap \cite{promptcap}         & Caption & Frozen GPT-3 (175B)        & 60.4 & 56.3 & 59.6 \\
    Prophet \cite{prophet}             & Caption + Tags & Frozen GPT-3 (175B) & 61.1 & 58.2 & 61.1 \\
    \midrule 
    InstructBLIP (7B) \cite{instruct_blip}  & Feature & Pretrain                   & 62.1 & 64.0 & 62.1 \\
    + \textit{Q\&A Prompts}             & Feature & Pretrain           & \textbf{64.3} (\textcolor{teal}{+2.2}) & \textbf{69.4} (\textcolor{teal}{+5.4}) & \textbf{68.1} (\textcolor{teal}{+6.0})\\
    \bottomrule 
  \end{tabular}
  \caption{Results of A-OKVQA \cite{aokvqa} and OK-VQA \cite{okvqa} comparing to standard baselines show that our method achieves state-of-the-art performance, outperforming previous methods by a large margin. The \textbf{best} results are highlighted.}
  \label{tab:results}
  }
\end{table}

\subsection{Implementation Details}
\label{details}
We use InstructBLIP \cite{instruct_blip} as our default MLLM, to both generate Q\&A pairs and perform answer predictions. 
RAM \cite{ram} is utilized as the image tagging model due to its strong recognition ability. The number of Q\&A pairs is set to $P=8$. To configure the visual-aware prompting module, we use the text encoder in CLIP \cite{clip} as our prompt encoder, and a value of 32 of the $k$ for the number of prompt embeddings $\textbf{F}_p\in \mathbb{R}^{k\times d_q}$, the same as the number of visual tokens $\textbf{F}_v\in \mathbb{R}^{k\times d_q}$. We train the models using AdamW \cite{adamw} as the optimizer with a learning rate of $2e^{-5}$ and the strategy of mixed precision training. It is important to note that the parameters of the image encoder, the large language model, and the prompt encoder are kept frozen to maintain their pre-trained representations.

\begin{wraptable}{r}{0.45\textwidth}
\resizebox{\linewidth}{!}{
  \centering
  \scalebox{1}{
  \begin{tabular}{l|cc}
    \toprule
    Models & A-OKVQA (\%) & OKVQA (\%) \\
    \midrule
    MiniGPT-4        & 54.2 & 52.5 \\
    + \textit{Q\&A Prompts}    & 61.2 (\textcolor{teal}{+7.0}) & 55.9 (\textcolor{teal}{+3.4})\\
    \midrule
    LLaVA-1.5        & 66.9 & 63.3 \\
    + \textit{Q\&A Prompts}    & 72.4 (\textcolor{teal}{+5.5}) & 65.2 (\textcolor{teal}{+1.9})\\
    \midrule
    InstructBLIP (T5)   & 58.5 & 55.1\\
    + \textit{Q\&A Prompts}    & 64.2 (\textcolor{teal}{+5.7}) & 57.3 (\textcolor{teal}{+2.2})\\
    \bottomrule
  \end{tabular}
  }
  \caption{Comparison of performance with different MLLMs as baselines.}
  \label{tab:blip_results}
}
\end{wraptable}

\subsection{Main Results}
\label{results}
\textbf{Compariosn with SoTAs.} The results in \cref{tab:results} demonstrate the superiority of our proposed \textit{Q\&A Prompts}. We have a significant improvement over existing SoTA methods, achieving an accuracy of 69.4\% on the validation set and 68.1\% on the test set in A-OKVQA, and an accuracy of 64.3\% in OK-VQA. We also observe that \textit{Q\&A Prompts} achieves a greater improvement on A-OKVQA (+5.4\% and +6.0\%) compared to OK-VQA (+2.2\%). This can be explained that: different from questions in OK-VQA that heavily rely on external knowledge, the questions in A-OKVQA emphasize visually-grounded reasoning with a comprehensive understanding of the image with diverse visual clues, rather than external knowledge retrieval. This aligns with \textit{Q\&A Prompts}’s philosophy, that is, instead of relying on accessing any external knowledge base such as Wikipedia \cite{wikidata}, ConceptNet \cite{concepetnet}, or GPT3 \cite{gpt3}, our focus is on enhancing reasoning capabilities by discovering rich visual clues hidden in the image and model.

\noindent \textbf{Different baselines.} To further validate the effectiveness of our method, we conduct experiments to compare various MLLMs as baseline models, including MiniGPT-4 \cite{minigpt4}, LLaVA-1.5 \cite{llava} and InstructBLIP (FLAN-T5-XL). \cref{tab:blip_results} shows our \textit{Q\&A Prompts} method consistently improves performance on A-OKVQA by approximately 5.5\% to 7.0\% and OKVQA by 1.9\% to 3.4\%, demonstrating the generalizability of \textit{Q\&A Prompts} across different vision-language models.

\noindent \textbf{Qualitative results.} We present qualitative results in \cref{fig:qualitive}. These cases demonstrate scenarios where correct answers are only possible with Q\&A prompts. For instance, the important clues provided by Q\&A Prompts of the [{\ttfamily{mirror}}],  [{\ttfamily{constructor}}], [{\ttfamily{cloudy}}] and [{\ttfamily{bible}}], lead to the successful reasoning of correct answers [{\ttfamily{bathroom}}], [{\ttfamily{visibility}}], [{\ttfamily{raining}}], and [{\ttfamily{getting married}}] respectively. This showcases the valuable contribution of the Q\&A prompts in certain types of tasks, which can explicitly mine rich visual clues and diverse world knowledge hidden in the image to help the model reason out correct answers. 

\begin{figure*}[t]
\label{fig:qualitive}
  \centering
   \includegraphics[scale=0.24]{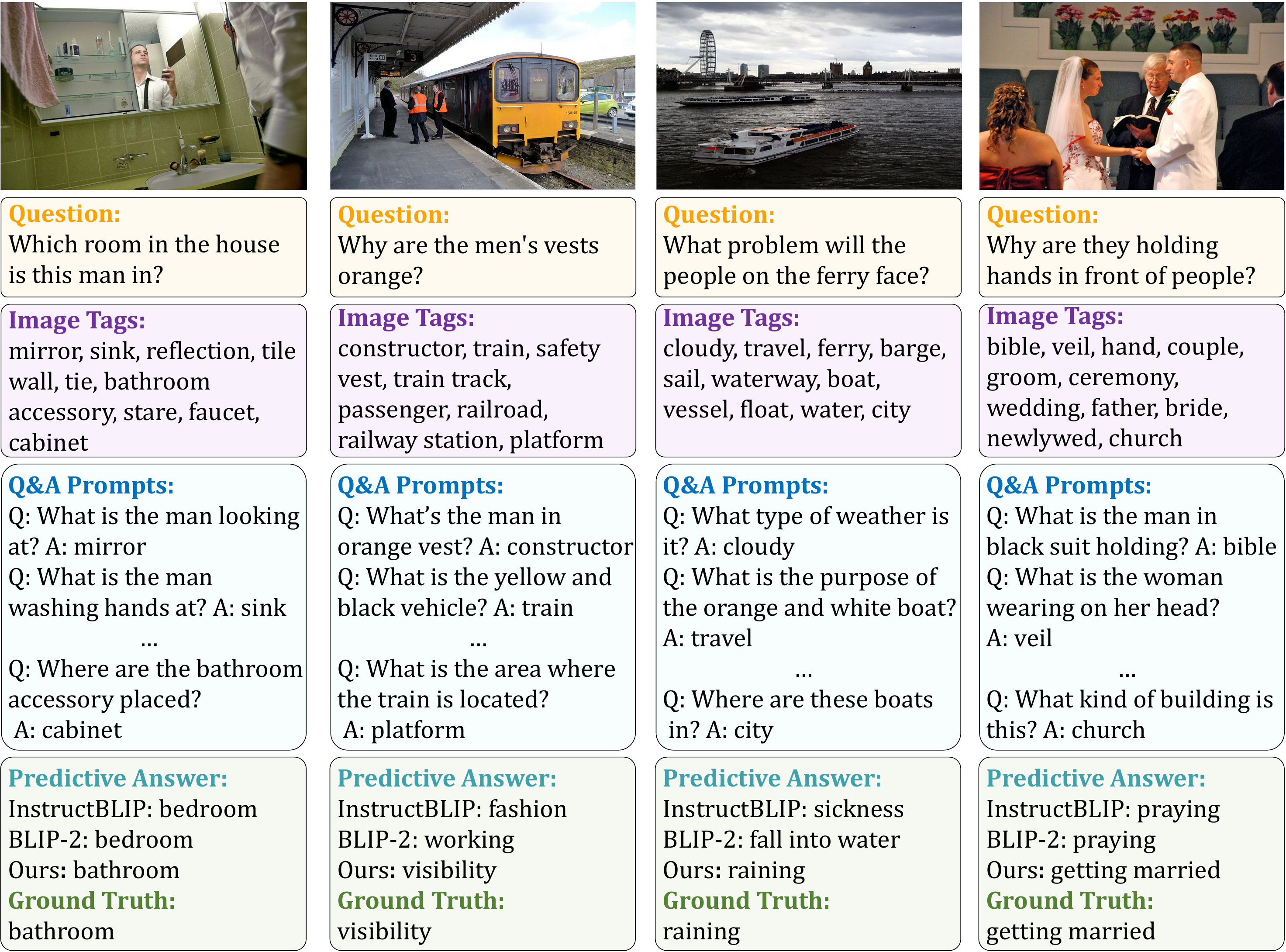}
   \caption{Representative cases with our \textit{Q\&A Prompts}. We denote the question, tags, Q\&A prompts, predictive answers, and ground truth respectively. Our \textit{Q\&A Prompts} can discover rich visual clues in the image and reason out the correct answer.}
   \label{fig:qualitive}
\end{figure*}

\begin{figure*}[t]
\label{fig:app3}
  \centering
   \includegraphics[scale=0.39]{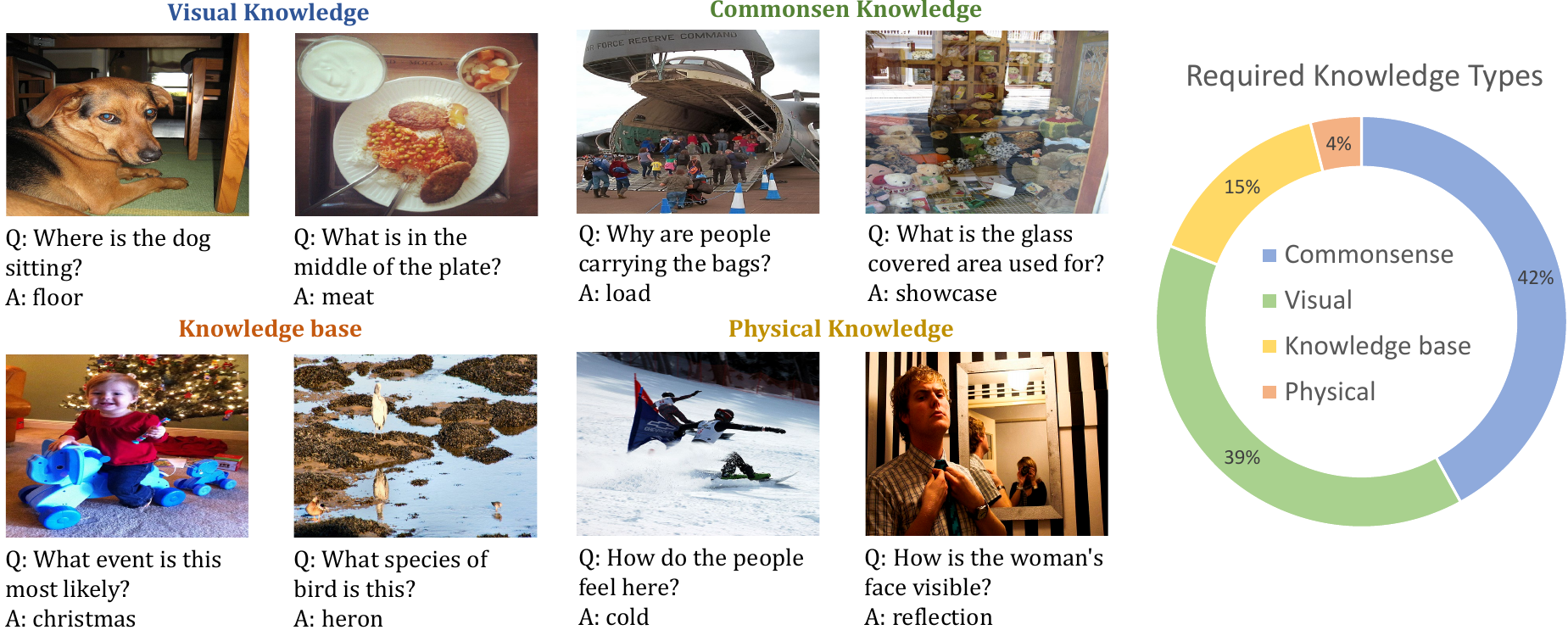}
   \caption{Generated Q\&A pairs with different knowledge types.}
   \label{fig:app3}
\end{figure*}

\begin{table}
\begin{floatrow}

\resizebox{\linewidth}{!}{

\capbtabbox{
  \scalebox{1}{
  \begin{tabular}{l|c}
    \toprule
    Q\&A source & Score \\
    \midrule
    Annotations in A-OKVQA    & 9.73\\
    VQG w/ training     & 8.86\\
    VQG w/o training    & 4.12\\
    \bottomrule
  \end{tabular}
  }
  \caption{The evaluation for reasonableness of different questions.}
  \label{tab:vqg_analysis}
}

\capbtabbox{
  \scalebox{0.7}{
  \begin{tabular}{lcc|cc}
    \toprule
     Q\&A \quad & Training Set \quad & Rouge-L \quad & A-OKVQA (\%) \quad & OK-VQA (\%)\\
    \midrule
    \XSolidBrush  &-               & -    & 64.0 & 62.1 \\
    \checkmark    &-               & -    & 61.9 & 61.0 \\
    \checkmark    &A-OKVQA         & 90.8 & \cellcolor{lightgray}{\textbf{69.4}} & 62.9 \\
    \checkmark    &OK-VQA          & 92.4 & 67.1 & \cellcolor{lightgray}{\textbf{64.3}} \\
    \checkmark    &Mixture         & 89.4 & 68.7 & 63.9 \\
    \bottomrule
  \end{tabular}
  }
  \caption{Ablation studies on training sets for VQG model training.}
  \label{tab:trainingset}
}

}
\end{floatrow}
\end{table}

\subsection{Analysis of the Generated Questions}
\label{ana_question}
To systematically evaluate the questions, we leverage GPT-4V to measure their reasonableness. Specifically, we randomly sample 250 <image, question, answer> tuples from the QA pairs generated by our VQG model, and ask GPT-4V to return an integer score of reasonableness ranging from 1 to 10 for them. For comparison, we also evaluate the scores of the manually annotated tuples in the A-OKVQA and those generated by directly prompting MLLM without training (with the same number of 250 tuples). \cref{tab:vqg_analysis} shows the questions generated by our VQG model (VQG w/ training) have a close quality to the manually annotated questions in A-OKVQA and are more reasonable than directly prompting a frozen MLLM (VQG w/o training). See appendix for more details. 

We also annotated what kind of world knowledge was required to answer the 250 sampled questions generated by our VQG model. The choices are the same with A-OKVQA including Commonsense Knowledge, Visual Knowledge, Knowledge Bases, and Physical Knowledge (explained in the appendix). The examples and distribution are shown in \cref{fig:app3}. Notably, sometimes there is no clear distinction between these categories and a question may belong to either.

\subsection{Ablation Study}
\label{ablation}
We investigate the role of our question-answer prompts and the visual-aware prompting module, based on InstructBLIP-Vicuna-7B and the A-OKVQA dataset.

\begin{table}
\begin{floatrow}
\resizebox{\linewidth}{!}{

\capbtabbox{
  \scalebox{1}{
  \begin{tabular}{@{}lc@{}}
    \toprule
    Fusion strategy & Accuracy (\%) \\
    \midrule
    Prepend   + Q\&A         & 66.6 \\
    \qquad \qquad  + Captions     & 64.2 \\
    \qquad \qquad  + Tags         & 64.7 \\
    VPM + Q\&A            & \cellcolor{lightgray}{\textbf{69.4}} \\
    \bottomrule
  \end{tabular}
}
  \caption{Ablation studies on strategies of fusing Q\&A pairs.}
  \label{tab:useqa}
}

\capbtabbox{
  \scalebox{1}{
  \begin{tabular}{@{}lc@{}}
    \toprule
    \#Num & Accuracy (\%) \\
    \midrule
    \qquad 1                          & 66.2 \\
    \qquad 4                          & 67.9 \\
    \qquad 8                          & \cellcolor{lightgray}{\textbf{69.4}} \\
    \qquad 16                         & 69.2 \\
    \bottomrule
  \end{tabular}
  }
  \caption{Ablation studies on numbers of question-answer pairs.}
  \label{tab:numberqa}
}

\capbtabbox{
  \scalebox{1}{
  \begin{tabular}{@{}lc@{}}
    \toprule
    Variants & Accuracy (\%) \\
    \midrule
    VPM   &\cellcolor{lightgray}{\textbf{69.4}} \\
    \quad w/ BERT             & 68.6 \\
    \quad w/o Fusion          & 68.6 \\
    \quad w/o Decoder         & 68.1 \\
    \bottomrule
  \end{tabular}
  }
  \caption{Ablation studies on the different components in the visual-aware prompting module.}
  \label{tab:component}
}


}
\end{floatrow}
\end{table}

\noindent \textbf{Different training sets for generated questions.}
We investigate how the VQG model will influence the performance by directly prompting the MLLM without training, or using different training sets to train our VQG model, including the A-OKVQA, OK-VQA, and a mixture of them. We report the Rouge-L score for the generated questions. Results in \cref{tab:trainingset} show that performance declines severely when using Q\&A prompts generated by MLLMs without training. Besides, the best performance is only achieved when the corresponding dataset is utilized as the training set, and a mismatch or mixture of datasets could lead to a slightly weaker performance. This could be attributed to the domain gap in the question types of these datasets.

\noindent \textbf{Strategies of fusing question-answer pairs.} In \cref{tab:useqa}, we explore a simple approach where we directly prepend the Q\&A pairs as a natural language to the instruction (denoted as Prepend + Q\&A) without encoded by the visual-aware prompting module ($\operatorname{VPM}$). For comparison, we also replace the question-answer pairs with the same number of image captions generated by BLIP-2 \cite{blip2} and tags recognized by RAM (denoted as Prepend + Captions and Prepend + Tags). The results in \cref{tab:useqa} show that the incorporation of question-answer prompts is more effective than pure captions and tags, and both directly prepending the question-answer pairs and using the $\operatorname{VPM}$ to encode them can improve performance. Notably, the models with $\operatorname{VPM}$ perform much better, which we attributed to its ability to combat some noise in the generated question-answer pairs and avoid interference with the target question.

\noindent \textbf{The number of the question-answer prompts.}
We also study to figure out the effect of using different numbers of question-answer pairs to construct the Q\&A Prompts. The results are displayed in \cref{tab:numberqa}. We denote the desired number as $P$. It can be observed that when $P$ is set to 8, the model achieves optimal performance. We can deduce that when $P$ is too large, it may introduce noisy pairs, while a too small $P$ can result in an ignorance of many essential and informative question-answer pairs.

\noindent \textbf{Components in visual-aware prompting module.} We validate the key components of the visual-aware prompting module in \cref{tab:component}. We first study the effectiveness of using different language models as the prompt encoder, where we replace the text encoder in CLIP with BERT \cite{bert} (w/ BERT), which results in a performance drop. This result is consistent with previous findings \cite{clip_benefit, LiMBeR}, which suggest that models initialized with CLIP \cite{clip} are more suitable for vision-language tasks due to their shared understanding of the visual and textual content. We then remove the visual gated fusion (w/o fusion), which also induces severe performance decline. This is expected since the unimodal language representations tend to involve ambiguity and bias. The additional incorporation of visual embeddings effectively allows for multi-modal interaction beneficial for vision-language reasoning. We also evaluate the importance of the resampler decoder (w/o decoder), where we directly feed the long sequence $\textbf{F}_{m}\in \mathbb{R}^{L_{s}\times d_{q}}$ into the large language model in \cref{eq:llm_decoder_a} and also caused a performance drop. The sequence length of target question $L_{q}$ is $\times 10$ times smaller than the sequence length $L_{s}$ of prompts $\mathcal{S}$ ($L_{q}\leq 10$, $L_{s}\geq 100$ here), resulting in a distraction to the understanding of the target question. This highlights the importance of resampling $\textbf{F}_{m}\in \mathbb{R}^{L_{s}\times d_{q}}$ into a shorter $\textbf{F}_{p}\in \mathbb{R}^{k\times d_{q}}$ ($k=32$ here).


\begin{figure}[t]
\label{fig:fig_qa}
  \centering
   \includegraphics[scale=0.38]{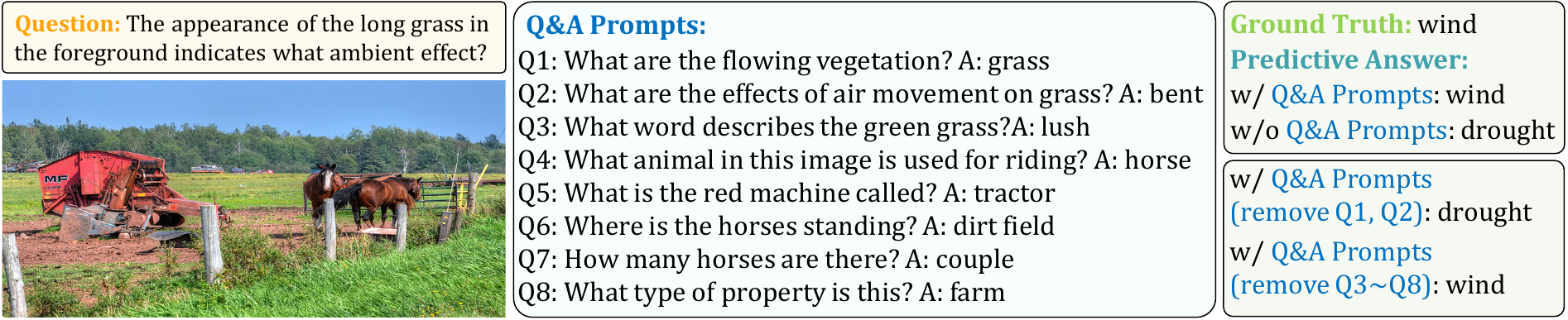}
   \caption{Qualitative analysis on how Q\&A Prompts work.}
   \label{fig:fig_qa}
\end{figure}

\section{A Further Qualitative Analysis of Q\&A Prompts}
\cref{fig:fig_qa} further reveals why Q\&A Prompts are necessary and how Q\&A Prompts work in particular case. To find out [{\ttfamily{what ambient effect}}] causes the appearance of grass, the key point is to grasp the movement of them. The Q\&A pairs $Q_1$ and $Q_2$ precisely capture the visual clues that [{\ttfamily{the grass are flowing}}] and [{\ttfamily{air movement cause the grass bent}}], which are beneficial for reasoning out the answer [{\ttfamily{wind}}]. We also test scenarios where we remove $Q_1$ and $Q_2$ or only reserve them, leading to the incorrect answer [{\ttfamily{drought}}] and correct answer [{\ttfamily{wind}}] respectively. This indicates that the quality of Q\&A Prompts is crucial, particularly when the question needs multiple reasoning steps beyond perception, and Q\&A Prompts serve as the intermediate steps that fill the logical gap.

\section{Conclusions and Limitations}
\label{sec:conclusion}
This paper has studied the problem of VQA tasks requiring reasoning over diverse world knowledge. We introduce a novel framework, \textit{Q\&A Prompts}, which effectively generates a set of question-answer prompts and encodes them with a visual-aware prompting module, significantly boosting the reasoning ability of current multi-modal large language models. We conducted a series of experiments on the A-OKVQA and OK-VQA benchmarks and achieved significant results compared with previous methods. Extensive ablations and comprehensive analyses have demonstrated the effectiveness and superiority of our method. 

\noindent \textbf{Limitations.}
While our method shows promising results, it is essential to acknowledge the potential presence of biases in the data as observed in previous VQA studies \cite{vqa_biases, cp_vqa}, as well as the lack of the ability for fine-grained counting and Optical Character Recognition. In our future work, we plan to mitigate these biases and weaknesses to enhance the reasoning ability of current models further.

%
%
\bibliographystyle{splncs04}
\bibliography{main}

\end{document}